\documentclass{llncs}
\usepackage{algorithm}
\usepackage{algorithmic}
\usepackage{graphicx}
\begin{document}
\title{Using Database Rule for Weak Supervised \newline Text-to-SQL Generation}
%
%
\author{
Tong Guo\inst{1}
Huilin Gao\inst{2}
}
%

%
\institute{Rokid AI Lab\and
China Electronic Technology Group Corporation Information Science Academy, Beijing, China}
\maketitle              
\begin{abstract}
We present a simple way to do the task of text-to-SQL problem with weak supervision. We call it Rule-SQL. Given the question and the answer from the database table without the SQL logic form, Rule-SQL use the rules based on table column names and question string for the SQL exploration first and then use the explored SQL for supervised training. We design several rules for reducing the exploration search space. For the deep model, we leverage BERT for the representation layer and separate the model to SELECT, AGG and WHERE parts. The experiment result on WikiSQL outperforms the strong baseline of full supervision and is comparable to the start-of-the-art weak supervised mothods.

\end{abstract}
\section{Introduction}

Semantic parsing is the tasks of translating natural language to logic form. Mapping from natural language to SQL (NL2SQL) is an important semantic parsing task for question answering system. But one key problem for semantic parsing is the hard label work. It is hard for human to write a lot of logic form corresponding to the question. 

On the other side, deep neural networks have achieved impressive performance on a lot of tasks such as image classification\cite{ref_proc1} and machine translation\cite{ref_proc10}. Researchers start to solve the NL2SQL problem by deep neural networks. BERT\cite{ref_proc2}, or Bidirectional Encoder Representations from Transformers\cite{ref_proc11}, is a new method of pre-training language representations which obtains state-of-the-art results on a wide array of Natural Language Processing (NLP) tasks. We employ BERT for the representation layer.

In order to solve the hard label problem of NL2SQL, we propose our simple weak supervised methods. Our key contribution are two folds:

1. We use rules to guide the SQL exploration, given the question, gold answer and the table header. The explored SQL can directly be used to supervised training for the NL2SQL task.

2. We design the deep neural model for the supervised training based on BERT and finish the experiments on WikiSQL. Based on the WikiSQL SQL structure, we design three model parts for the SELECT, AGG and WHERE clause. The code is available. \footnote[1]{\url{https://github.com/guotong1988/Rule-SQL}}

\begin{figure}
\centering
\includegraphics[width=\textwidth]{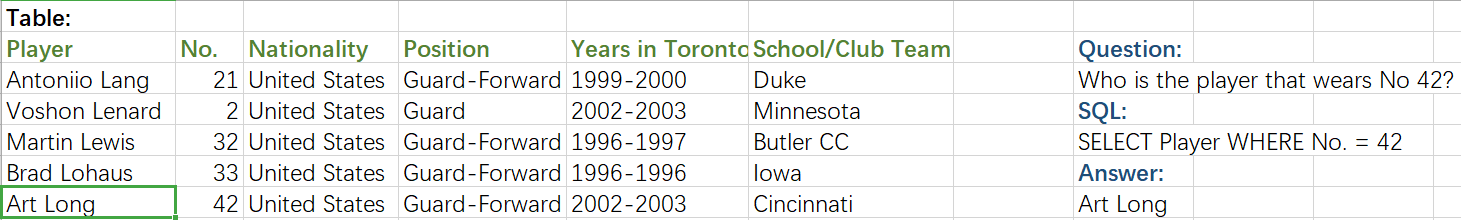}
\caption{An example of WikiSQL dataset} \label{fig1}
\end{figure}

\section{Related Work}

WikiSQL \cite{ref_proc3} is a large semantic parsing dataset. It has 80654 natural language and corresponding SQL pairs. The examples of WikiSQL are shown in fig. 1.

For full supervised training methods, there were a lot of work: \cite{ref_proc3} proposes Seq2sql which separates the SQL into three sub-part to solve and outperforms the sequence-to-sequence baseline.  \cite{ref_proc5} proposes SQLNet which employ the sequence-to-set and attention technique to solve the order problem of WHERE clause.  \cite{ref_proc6} propose TypeSQL which take the additional knowledge information as input.  \cite{ref_proc7} propose Coarse2Fine model which first generates raw output and then refine the raw output to generate a better result. 

On the other hand, there were some works that solve this problem by weak supervised methods, which means the model do not use the SQL or logic form for training. Neural Symbolic Machines(NSM) \cite{ref_proc8} develop a reinforcement framework and use the execution reward to guide the training. They further do a lot of improvement for the NSM reinforcement framework \cite{ref_proc9} and achieve state-of-the-art result. Our methods do not conflict to NSM.

\section{Database Rule Guided SQL Exploration}

\begin{algorithm}
\caption{SQL Exploration}
\label{alg:A}
\begin{algorithmic}
\STATE{$explored\_SQL$ = set()}
\FOR{each $question \in all\_question$}
\STATE{$SQL\_to\_try$ = None}
\WHILE{\TRUE}
\STATE{$SQL\_to\_try$ = do\_greedy\_explore($table\_header, question$)}
\STATE{$answer$ = execute($SQL\_to\_try$)}
\IF{$answer == gold\_answer$}
\IF{check\_rules($SQL\_to\_try$)==\TRUE}
\STATE{break}
\ENDIF
\ENDIF
\ENDWHILE            
\STATE{$explored\_SQL$.add($SQL\_to\_try$)}
\ENDFOR
\end{algorithmic}
\end{algorithm}

In this section we describe our SQL exploration based on the database design rules and the question string.

We first use greedy search to construct full SQL based on the WikiSQL dataset. For example, if the table has 5 columns, we should try at most 5 times of SQL execution for the SELECT column slot. The agg operation has 6 possible choices: (empty), AVERAGE, SUM, COUNT, MIN, MAX. So we should try at most 6 times of SQL execution for the SELECT agg slot. The total amount we should try for SELECT column slot and SELECT agg slot is 5*6=30 times.

For reducing the search space for the greedy exploration, we use the methods below: If the agg slot is empty and the where condition number is more than 1, the result of the SQL with 1 condition should contains the SQL with 2 condition. For example, the result of (SELECT * FROM WHERE condition1) should contains the result of (SELECT * FROM WHERE condition1 and condition2). So after we have searched the result for the first condition, we could fix the first condition and search for the second condition.

We then use exact full answer match between the constructed SQL's answer and gold SQL's answer. Then we use the rules below to handle the different explored SQLs with the same answer.

The rules based on the database table and question string is to remove the duplicate SQLs with the same answer. We describe the rules based on database for the problem of different SQLs with same answer as below:

1. If agg operation is empty, the answer should at the same column of explored SELECT column. For example, The execution result of (SELECT column1 FROM table1) should contains the answer. By this rule, we can verify the rightness of SELECT column exploration.

2. if agg operation is empty and the SELECT slot is set to *, the answer should be contained in the rows of SQL execution result.  For example, the execution result of (SELECT * FROM table1 WHERE ...) should contains the answer. By this rule, we can verify the raw rightness of WHERE clause.

3. The question string should contains the same words which appears in the SQL WHERE clause. For example: The result of (SELECT where-clause1-column FROM table1) should contains some words in the question string.

4. If the operation slot of WHERE is EQUAL, then the gold answer and value slot of WHERE should at the same row.

5. If the WHERE value slot is a string, then the WHERE op slot must be EQUAL. 

6. If the gold answer is a string, then the agg slot should be empty.

7. If the gold answer is a number which do not appear in any place of the table data, the agg should not be empty and can only be COUNT or SUM or AVERAGE.  

The explored SQL are used for the full supervised training. The full algorithm is shown in Algorithm 1:

\begin{figure}
\centering
\includegraphics[width=\textwidth]{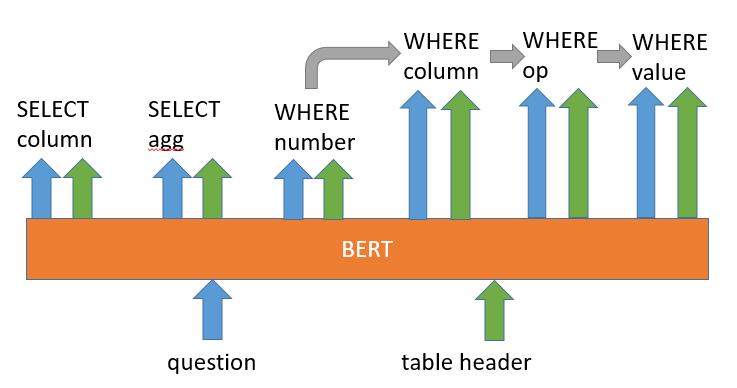}
\caption{The deep model} \label{fig1}
\end{figure}

\section{The Deep Neural Model}

Based on the Wikisql dataset, we also use three sub-model to predict the SELECT part, AGG part and WHERE part. The whole model is shown in fig. 2.

We use BERT as the representation layer. The question and table header are concat and then input to BERT, so that the question and table header have the attention interaction information of each other.
We denote the BERT output of question and table header as $Q$ and $H$

\subsection{SELECT column}
Our goal is to predict the column name in the table header. The inputs are $Q$ and $H$. The output are the probability of SELECT column: 
\begin{equation}
P(sc|Q,H)
\end{equation}

\subsection{SELECT agg}
Our goal is to predict the agg slot. The inputs are $Q$ and the output are the probability of SELECT agg:
\begin{equation}
P(sa|Q) 
\end{equation}

\subsection{WHERE number}

Our goal is to predict the where number slot. The inputs are $Q$ and $H$. The output are the probability of WHERE number:
\begin{equation}
P(wn|Q,H) 
\end{equation}

\subsection{WHERE column}

Our goal is to predict the where column slot for each condition of WHERE clause. The inputs are $Q$, $H$ and $P_{wn}$. The output are the top $wherenumber$ probability of WHERE column:
\begin{equation}
P(wc|Q,H,P_{wn}) 
\end{equation}

\subsection{WHERE op}

Our goal is to predict the where column slot for each condition of WHERE clause. The inputs are $Q$, $H$, $P_{wc}$ and $P_{wn}$. The output are the probability of WHERE op slot:
\begin{equation}
P(wo|Q,H,P_{wn},P_{wc}) 
\end{equation}

\subsection{WHERE value}
Our goal is to predict the where column slot for each condition of WHERE clause. The inputs are $Q$, $H$, $P_{wn}$, $P_{wc}$ and $P_{wo}$. The output are the probability of WHERE value slot:
\begin{equation}
P(wv|Q,H,P_{wn},P_{wc},P_{wo}) 
\end{equation}

\section{Experiments}

In this section, we present more details of the model and the evaluation on the dataset. Pre-trained BERT models (BERT-Base-Uncased) are loaded and fine-tuned with Adam optimizer with learning rate $1*10^-5$. The rest parts of the whole model's learning rate is $1*10^-3$. The batch size is 8. We use the origin BERT tokenizer with the same vocabulary of BERT-Base-Uncased and process all the token to lower case. Our neural network model is implemented in Pytorch.

\subsection{Experiment result}

In this section, we evaluate Rule-SQL versus other approachs on the WikiSQL dataset.
The explored SQL does not work well for where clause which condition number is more than 1. See Table 1. for detail. See Table 2. for the result that compared to other full supervised methods and Table 3. for the result that compared to other weak supervised methods.

\begin{table}
\caption{Remove the where condition which is more than 1 or 2}\label{tab1}
\centering
\begin{tabular}{|l|l|l|l|l|}
\hline
Model & Logic Form Dev Acc & Execution Dev Acc & Logic Form Test Acc & Execution Test Acc\\
\hline
Rule-SQL 1-4 condition & 47.9$\%$ & 61.1$\%$ & 47.5$\%$ & 61.0$\%$ \\ 
\hline
Rule-SQL 1-2 condition & 51.1$\%$ & 63.2$\%$ & 49.0$\%$ & 63.0$\%$ \\ 
\hline
Rule-SQL 1 condition & 68.7$\%$ & 81.2$\%$ & 68.5$\%$ & 81.0$\%$ \\ 
\hline
\end{tabular}
\end{table}

\begin{table}
\caption{Compare Rule-SQL to other full supervised models}\label{tab1}
\centering
\begin{tabular}{|l|l|l|l|l|}
\hline
Model & Logic Form Dev Acc & Execution Dev Acc & Logic Form Test Acc & Execution Test Acc\\
\hline
Seq2seq & 23.3$\%$ & 37.0$\%$ & 23.4$\%$ & 35.9$\%$ \\ 
\hline
Seq2SQL & 49.5$\%$ & 60.8$\%$ & 48.3$\%$ & 59.4$\%$ \\
\hline
SQLNet & - & 69.8$\%$ & - & 68.0$\%$ \\
\hline
Rule-SQL 1 condition & 68.7$\%$ & 81.2$\%$ & 68.5$\%$ & 81.0$\%$ \\ 
\hline
\end{tabular}
\end{table}

\begin{table}
\caption{Compare Rule-SQL to other weak supervised models}\label{tab1}
\centering
\begin{tabular}{|l|l|l|l|l|}
\hline
Model & Logic Form Dev Acc & Execution Dev Acc & Logic Form Test Acc & Execution Test Acc\\
\hline
NSM & - & 72.2$\%$ & - & 72.1$\%$ \\
\hline
Rule-SQL 1 condition & 68.7$\%$ & 81.2$\%$ & 68.5$\%$ & 81.0$\%$ \\ 
\hline
\end{tabular}
\end{table}

\subsection{Badcase}
In this section we display the badcase of SQLs with the same answer.

\vspace{2\baselineskip}

Gold SQL 1:

\textbf{SELECT column1 WHERE condition1 and condition2}

Explored SQL 1:

\textbf{SELECT column1 WHERE condition1}

In this case, condition1 could lead to the gold answer without condition2. For example, condition1 result in the same row to condition2.

\vspace{2\baselineskip}

Gold SQL 2:

\textbf{SELECT MAX(column1) WHERE condition1}

Explored SQL 2:

\textbf{SELECT MIN(column1) WHERE condition1}

In this case, condition1 only contains 1 row. So MAX and MIN lead to the same result.

\section{Conclusion}
In this paper, in order to solve the hard label problem for semantic parsing, we introduce our idea that leveraging the database rules to guide the SQL exploration, given only question answer pairs. The explored SQL can be directly used for full supervised training. Our methods does not conflict to other weak supervised methods\cite{ref_proc9}. We employ BERT-based model to predict the three parts that correspond to the aggregation operator, the SELECT column, and the WHERE clause. The experiment result outperforms full supervised strong baseline and is comparable to state-of-the-art weak supervised methods. We will add the execution guided decoding\cite{ref_proc4} experiment result for the future work. And we will try to approve the experiments result in the future by borrowing idea from \cite{ref_proc12}.

\end{document}